\documentclass[11pt]{article}

\usepackage[margin=1in]{geometry}
\usepackage{amsmath,amssymb,amsfonts}
\usepackage{graphicx}
\usepackage{booktabs}
\usepackage{hyperref}
\usepackage{cleveref}
\usepackage{xcolor}
\usepackage{microtype}
\usepackage{natbib}
\usepackage{caption}
\usepackage{subcaption}
\usepackage{multirow}
\usepackage{array}

\hypersetup{colorlinks=true, linkcolor=blue!60!black, citecolor=green!50!black, urlcolor=blue!70!black}

\newcommand{\DW}{\Delta W}

\newcommand{\kstar}{k^*}

\title{%
  \textbf{Global Low-Rank, Local Full-Rank:\\
  The Holographic Encoding of Learned Algorithms}%
}

\author{
  Yongzhong Xu\footnote{\texttt{abbyxu@gmail.com}. Code: \url{https://github.com/skydancerosel/holographic_encoding}}
}

\date{February 20, 2025}

\begin{document}
\maketitle

\begin{abstract}
Grokking---the abrupt transition from memorization to generalization after extended training---has been linked to the emergence of structured, low-dimensional representations.
Yet neural network parameters live in extremely high-dimensional spaces.
How can a low-dimensional learning process produce solutions that resist low-dimensional compression?

We investigate this question in multi-task modular arithmetic, training shared-trunk Transformers with separate heads for addition, multiplication, and a quadratic operation modulo~97.
Across three model scales (315K--2.2M parameters) and five weight decay settings, we systematically compare three reconstruction methods: per-matrix SVD, joint (cross-matrix) SVD, and trajectory PCA.

Across all conditions, we find that grokking trajectories are confined to a 2--6 dimensional global subspace, while individual weight matrices remain effectively full-rank.
Reconstruction from 3--5 trajectory PCs recovers $>$95\% accuracy, whereas both per-matrix and joint SVD fail catastrophically at sub-full rank.
Even when static decompositions capture most spectral energy, they destroy task-relevant structure.

These results show that learned algorithms are encoded through dynamically coordinated updates spanning all matrices, rather than through localized low-rank components.
We term this the \emph{holographic encoding principle}: grokked solutions are globally low-rank in the space of learning directions but locally full-rank in parameter space.
This duality has implications for compression, interpretability, and understanding how neural networks encode algorithmic structure.
\end{abstract}

\section{Introduction}
\label{sec:intro}

Neural networks trained on algorithmic tasks often exhibit a striking phenomenon: after a prolonged period of memorization, they abruptly transition to systematic generalization.
This delayed onset of generalization, known as \emph{grokking}, has been observed across a range of settings and architectures \citep{power2022grokking, davies2023unifying} and has been linked to the emergence of structured internal representations and implicit regularization effects \citep{liu2022omnigrok, merrill2023tale}.

Recent work has shown that grokking is associated with the discovery of interpretable circuits, such as Fourier-based representations for modular arithmetic \citep{nanda2023progress, zhong2023clock}, and that weight decay plays a central role in driving the transition from memorization to algorithmic solutions \citep{varma2023explaining}.
At the same time, analyses of training dynamics have revealed that grokking trajectories are often confined to remarkably low-dimensional subspaces \citep{lyu2023dichotomy, xu2026manifolds, xu2026curved}, suggesting that the effective complexity of learning is far lower than the ambient parameter dimension.

These observations give rise to a fundamental puzzle.
On the one hand, the learning process appears highly constrained: projecting the training trajectory onto a small number of principal directions is sufficient to reconstruct a near-optimal model.
On the other hand, the resulting parameters resist compression: truncating the singular value decomposition of even a single weight matrix destroys performance.
How can a learning process that is globally low-dimensional produce solutions that are locally irreducible?

The multi-task setting sharpens this puzzle further \citep{xu2026geometry}.
When a shared model jointly learns multiple algorithmic operations, it must simultaneously represent distinct computational circuits within a common parameterization.
Using gradient-based probes, we find that these circuits are cleanly separable in activation space, with nearly orthogonal task-specific subspaces.
Yet this separation is invisible in the weights: every major parameter matrix contributes to all tasks, and no static factorization isolates task-specific structure.

In this work, we resolve this apparent contradiction by identifying a global--local duality in grokked solutions.
We show that learning unfolds along a low-dimensional invariant subspace of coordinated parameter updates, while the final implementation of the learned algorithms is distributed across high-dimensional, full-rank weight matrices.
The information required for computation is not stored in localized components, but in dynamically structured correlations that emerge over the course of training.

We term this phenomenon the \textbf{holographic encoding principle}: learned algorithms are \emph{globally low-rank} in the space of learning directions but \emph{locally full-rank} in parameter space.
This duality explains why static compression methods fail, why trajectory-informed decompositions succeed, and why algorithmic circuits appear separable in activation space but entangled in weight space.

We demonstrate this through a systematic analysis across 15 experimental conditions (3 model sizes $\times$ 5 weight decay values):
\begin{enumerate}
    \item \textbf{Circuit separability}: Gradient-based directional ablation reveals that task circuits are nearly orthogonal in activation space (mean selectivity index $> 0.98$, subspace overlap $< 0.08$), confirming that the shared trunk implements separable computations for each task.
    \item \textbf{Trajectory PCA}: Projecting the final weight change onto 3--5 principal components of the training trajectory recovers $>$95\% accuracy. The learning dynamics are confined to a low-dimensional invariant subspace.
    \item \textbf{Per-matrix SVD}: Independently truncating each weight matrix's change ($\DW$) to rank 64 yields chance-level performance (mean accuracy $\approx 1\%$). Individual matrices are effectively full-rank for task purposes.
    \item \textbf{Joint SVD}: Stacking all weight-change vectors and performing a single SVD also fails at sub-full rank. Cross-matrix correlations exist but cannot be discovered without the training trajectory.
\end{enumerate}

The key insight is that the cross-matrix correlations encoding the solution are \emph{dynamically structured}---they can be captured by the learning trajectory but not by any static decomposition of the final weights alone.
Circuits are separable in the space of computations but holographically entangled in the space of parameters.

\section{Experimental Setup}
\label{sec:setup}

\subsection{Tasks and Data}

We study multi-task modular arithmetic modulo $P = 97$.
Each model is trained on three tasks simultaneously:
\begin{itemize}
    \item \textbf{Addition}: $(x, y) \mapsto (x + y) \bmod 97$
    \item \textbf{Multiplication}: $(x, y) \mapsto (x \cdot y) \bmod 97$
    \item \textbf{Quadratic}: $(x, y) \mapsto (x^2 + y^2) \bmod 97$
\end{itemize}
The input space consists of all $97^2 = 9{,}409$ pairs, split 50/50 into training and test sets.
The three tasks share a common transformer trunk but have separate linear classification heads, each mapping to 97 output classes.

\subsection{Model Architecture}

We train three model scales, summarized in \Cref{tab:models}.
All models use a shared token embedding ($P \to d_\text{model}$), a learned 2-position positional embedding, a pre-norm transformer encoder trunk, a final layer norm, and three independent linear heads for the three tasks.
We use the mean of the two token representations as input to each head.

\begin{table}[h]
\centering
\caption{Model configurations. All models use pre-norm transformer encoders with $d_\text{ff} = 2 \times d_\text{model}$ and no dropout.}
\label{tab:models}
\begin{tabular}{lccccr}
\toprule
\textbf{Model} & $d_\text{model}$ & Layers & Heads & $d_\text{ff}$ & \textbf{Parameters} \\
\midrule
Baseline & 128 & 2 & 4 & 256 & 315,427 \\
Medium   & 128 & 4 & 4 & 256 & 580,387 \\
Large    & 256 & 4 & 8 & 512 & 2,209,059 \\
\bottomrule
\end{tabular}
\end{table}

\subsection{Training}

All models are trained with AdamW (learning rate $10^{-3}$, $\beta = (0.9, 0.98)$) with full-batch gradient descent.
We sweep weight decay $\lambda \in \{0.1, 0.2, 0.3, 0.5, 1.0\}$ across all three scales, yielding 15 experimental conditions.
Training continues until all three tasks grok (test accuracy $> 95\%$) or until a maximum step count is reached.
All 15 conditions achieve full grokking (mean test accuracy $> 97\%$).

Higher weight decay dramatically accelerates grokking: the baseline model groks at ${\sim}157$K steps with $\lambda = 0.1$ versus ${\sim}25$K steps with $\lambda = 1.0$.
The large model with $\lambda = 1.0$ groks by step 6,400.
A detailed analysis of the training dynamics, phase structure, and transverse instabilities of these models is given in \citet{xu2026geometry}.

\subsection{Analysis Methods}

We compare three methods for reconstructing the final model from compressed representations of the weight change $\DW = W_\text{final} - W_\text{init}$:

\paragraph{Trajectory PCA.}
We collect model checkpoints throughout training, flatten each into a parameter vector, subtract the initial parameters, and compute the SVD of the resulting trajectory matrix.
To reconstruct at rank $k$, we project $\DW_\text{final}$ onto the top-$k$ principal components and evaluate the reconstructed model.

\paragraph{Per-matrix SVD.}
For each weight matrix independently, we compute $\DW_i = W_{\text{final},i} - W_{\text{init},i}$ and replace it with its rank-$k$ SVD approximation $U_{:,:k} \Sigma_{:k} V_{:k,:}^\top$.
All matrices are truncated simultaneously at the same rank.

\paragraph{Joint SVD.}
We flatten each trunk $\DW_i$ into a vector, stack all $n$ vectors as rows of an $n \times d_\text{max}$ matrix (zero-padded), and compute the SVD of this stacked matrix.
Reconstruction at rank $k$ captures cross-matrix correlations that per-matrix SVD misses.

In all trunk-only experiments, the task heads and token embeddings are kept at their final trained values; only the transformer encoder matrices are reconstructed.

\section{Results}
\label{sec:results}

\subsection{Learning Trajectories Are Low-Dimensional}
\label{sec:global}

We begin by establishing the global structure: despite the high dimensionality of parameter space, the grokking trajectory is confined to a remarkably low-dimensional subspace.

\Cref{tab:kstar} reports the minimum number of trajectory PCs required to reconstruct the final model above 95\% and 99\% of baseline accuracy, across all 15 conditions.
The results are striking: $\kstar(95\%) = 2$--$6$ across all model sizes and weight decays.
Even the largest model (2.2M parameters) can be reconstructed from just 2--4 trajectory PCs at the 95\% threshold.

\begin{table}[h]
\centering
\caption{Trajectory PC $\kstar$: minimum PCs for reconstruction above 95\% / 99\% of baseline accuracy. All three tasks must individually pass the threshold.}
\label{tab:kstar}
\begin{tabular}{l ccccc ccccc}
\toprule
& \multicolumn{5}{c}{$\kstar(95\%)$} & \multicolumn{5}{c}{$\kstar(99\%)$} \\
\cmidrule(lr){2-6} \cmidrule(lr){7-11}
\textbf{Model} & $\lambda{=}0.1$ & $0.2$ & $0.3$ & $0.5$ & $1.0$ & $\lambda{=}0.1$ & $0.2$ & $0.3$ & $0.5$ & $1.0$ \\
\midrule
Baseline & 5 & 5 & 6 & 3 & 5 & 7 & 16 & 11 & 5 & 7 \\
Medium   & 5 & 5 & 4 & 3 & 4 & 15 & 7 & 5 & 5 & 5 \\
Large    & 4 & 3 & 3 & 2 & 2 & $>$30 & 6 & $>$30 & 3 & 3 \\
\bottomrule
\end{tabular}
\end{table}

\begin{figure}[t]
    \centering
    \includegraphics[width=\textwidth]{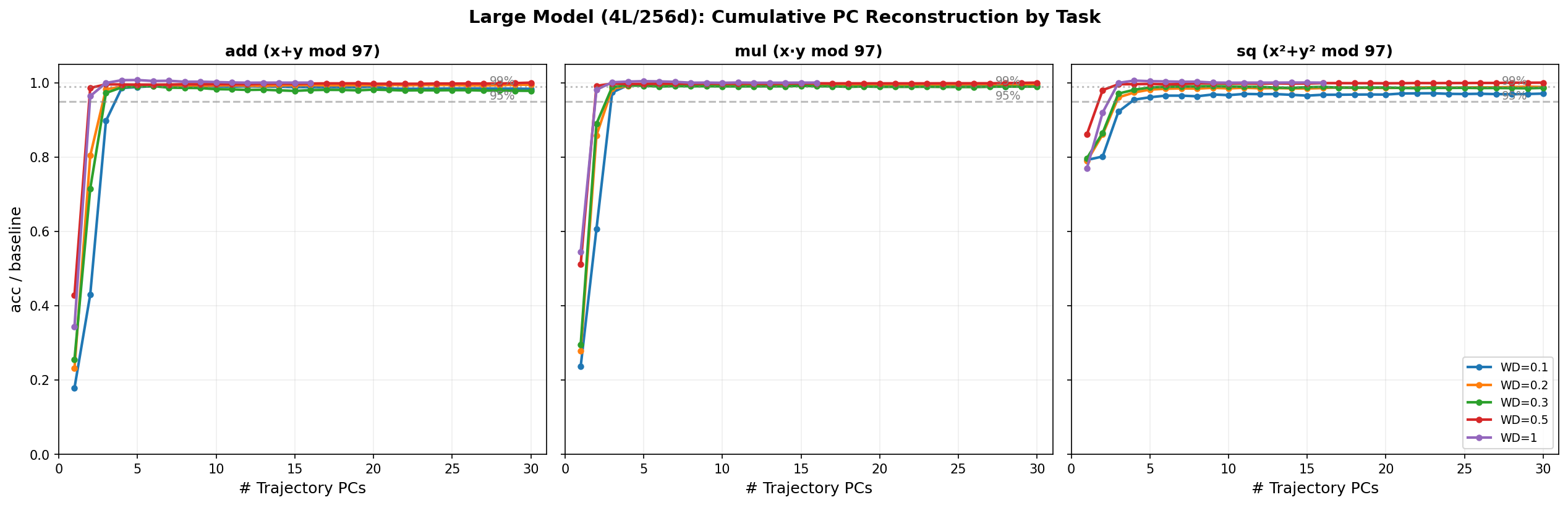}
    \caption{Cumulative trajectory PC reconstruction for the large model (4L/256d), by task. Addition and multiplication saturate by $k = 3$--$5$ PCs. The quadratic task exhibits a characteristic long tail at low weight decay, requiring more PCs for the final 2--3\% of accuracy.}
    \label{fig:traj_pca}
\end{figure}

\Cref{fig:traj_pca} shows the cumulative reconstruction curves.
At $k = 5$ PCs, the reconstructed model achieves 93--99\% of baseline accuracy across all conditions.
The first PC alone explains 90--97\% of the trajectory variance, and 3 PCs cumulatively explain $>$96\%.

At the 99\% threshold, some conditions---particularly the large model at low weight decay---require substantially more PCs ($>$30).
This long tail is driven primarily by the quadratic task ($x^2 + y^2 \bmod 97$), which requires fine-grained weight structure that is distributed across many small PCs.
Addition and multiplication saturate by $k = 3$--$5$, while the quadratic task plateaus at ${\sim}97\%$ of baseline and climbs slowly thereafter.

Crucially, all trajectory PCs are \emph{task-agnostic}: removing any single critical PC (1, 2, or 3) kills all three tasks simultaneously.
No PC is specific to a single task.
The trajectory PCs capture shared trunk infrastructure, not task-specific circuits.

Thus, despite extreme overparameterization, the grokking process unfolds along a highly constrained low-dimensional invariant subspace in parameter space.

\subsection{Individual Weight Matrices Are Effectively Full-Rank}
\label{sec:local}

We now show that the local structure of the solution is the opposite of its global structure: individual weight matrices are effectively full-rank for task purposes.

For each trunk weight matrix, we compute the SVD of $\DW_i$ and measure the rank required to capture 90\% and 99\% of the Frobenius norm.
Across all 15 conditions, the mean $k_{90}$ is 50--52\% of the full matrix dimension, and the mean $k_{99}$ is 77--80\% (\Cref{tab:svd_energy}).
The spectral structure of the weight changes is spread broadly---there is no sharp spectral gap.

\begin{table}[h]
\centering
\caption{Per-matrix SVD energy concentration: mean rank required to capture 90\% / 99\% of $\|\DW_i\|_F^2$ across trunk matrices, expressed as percentage of full rank.}
\label{tab:svd_energy}
\begin{tabular}{l ccccc}
\toprule
& \multicolumn{5}{c}{$k_{90}$ / $k_{99}$ (\% of full rank)} \\
\cmidrule(lr){2-6}
\textbf{Model} & $\lambda{=}0.1$ & $0.2$ & $0.3$ & $0.5$ & $1.0$ \\
\midrule
Baseline ($d{=}128$) & 50/79 & 50/79 & 51/80 & 51/80 & 50/80 \\
Medium ($d{=}128$)   & 51/80 & 51/80 & 52/80 & 52/80 & 52/80 \\
Large ($d{=}256$)    & 44/77 & 47/78 & 49/79 & 49/79 & 50/80 \\
\bottomrule
\end{tabular}
\end{table}

\begin{figure}[t]
    \centering
    \includegraphics[width=\textwidth]{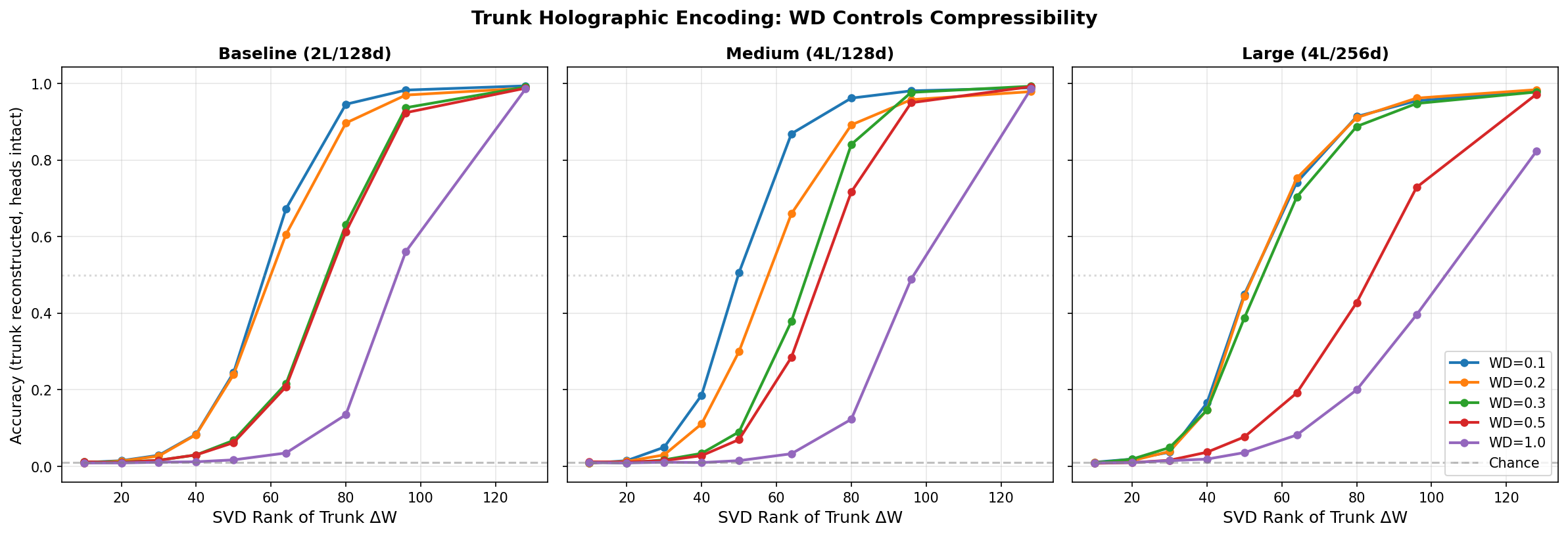}
    \caption{Trunk-only SVD reconstruction accuracy versus rank $k$, across weight decay and model size. Even at rank 64 (half of $d_\text{model}$ for baseline/medium), most conditions remain near chance. Recovery requires 80--100+ singular components per matrix.}
    \label{fig:permatrix}
\end{figure}

But energy concentration does not tell the full story.
When we reconstruct the model by truncating all trunk matrices to rank $k$ (keeping heads at their final values), performance remains at chance level even at rank 50---capturing ${\sim}88\%$ of the Frobenius energy---and only begins to recover around rank 80--100 (\Cref{fig:permatrix}).
At rank 64, the trunk-only reconstruction achieves:
\begin{itemize}
    \item Baseline model: 67\% (WD=0.1) to 3.5\% (WD=1.0)
    \item Large model: 74\% (WD=0.1) to 8.2\% (WD=1.0)
\end{itemize}

This reveals a critical disconnect: 88\% of the energy is captured, but the task-relevant information resides in the remaining 12\%.
The singular components that matter for function are distributed across the tail of the spectrum, not concentrated in the top singular values.
Thus, spectral energy concentration is not a reliable proxy for functional compression: components that are energetically small can be essential when coordinated across matrices.

Higher weight decay exacerbates the incompressibility---despite producing the same spectral shape (entropy $\approx 0.87$ across all conditions), stronger regularization pushes the task-critical information deeper into the spectral tail.

The final solution is not low-rank within individual weight matrices.

\subsection{Joint Static Low-Rank Factorization Also Fails}
\label{sec:joint}

The per-matrix SVD decomposes each $\DW_i$ independently, missing cross-matrix correlations.
Could a joint decomposition that captures these correlations succeed?

We test this by stacking all $n$ trunk $\DW$ vectors (flattened) as rows of a single $n \times d_\text{max}$ matrix and computing its SVD.
For the baseline model ($n = 12$ trunk matrices) and the medium/large models ($n = 24$), this provides a principled way to capture cross-matrix structure.

\Cref{tab:three_methods} summarizes the results.
Joint SVD at sub-full rank also fails to recover accuracy.
At half rank ($r = n/2$), all conditions remain near chance.
Only at full rank ($r = n$)---which is equivalent to lossless reconstruction---does accuracy recover.

The transition is sharp: for the large model at $\lambda = 0.1$, joint SVD accuracy jumps from 48\% at $r = 20$ to 99\% at $r = 24$ (full rank).
There is no graceful degradation; removing even a few components from the cross-matrix decomposition destroys the solution.

\begin{figure}[t]
    \centering
    \includegraphics[width=\textwidth]{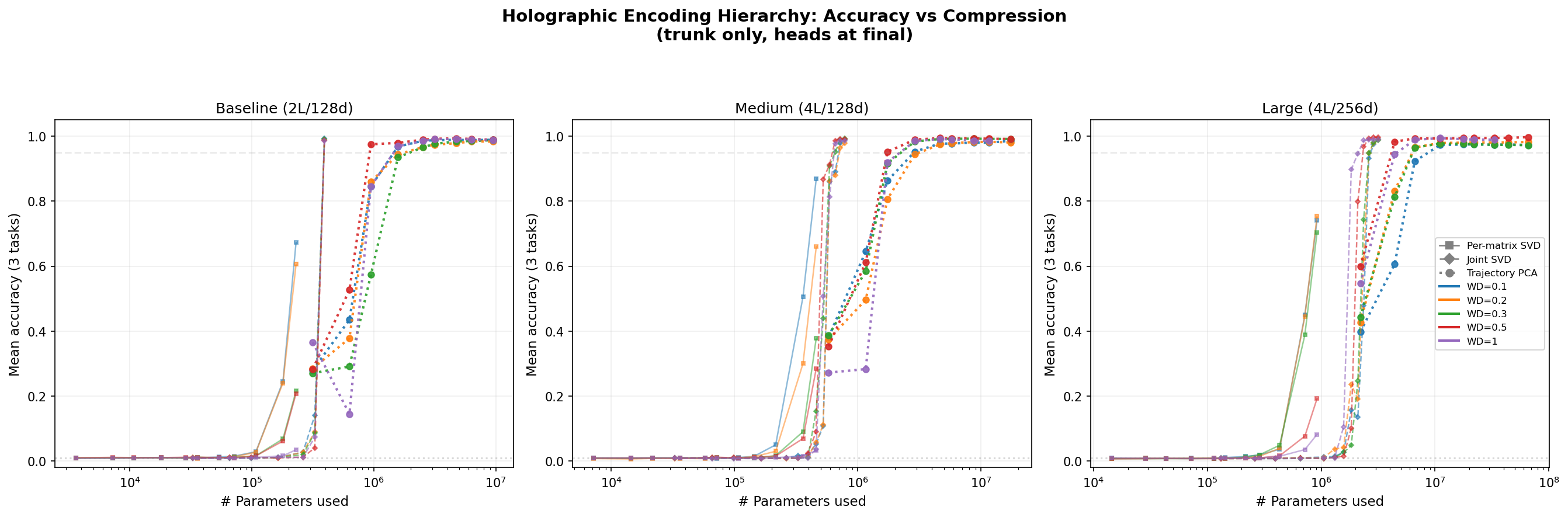}
    \caption{Accuracy versus number of parameters used for the three reconstruction methods, across all model sizes. Trajectory PCA (circles, dotted) achieves $>$95\% accuracy at a fraction of the parameters required by per-matrix SVD (squares, solid) or joint SVD (diamonds, dashed).}
    \label{fig:methods_comparison}
\end{figure}

\begin{table}[h]
\centering
\caption{Comparison of three reconstruction methods (mean accuracy across three tasks, trunk-only). PM = per-matrix SVD, Joint = joint SVD, Traj = trajectory PCA.}
\label{tab:three_methods}
\begin{tabular}{ll rrr rrr}
\toprule
& & \multicolumn{3}{c}{\textbf{Static decomposition}} & \multicolumn{3}{c}{\textbf{Trajectory-informed}} \\
\cmidrule(lr){3-5} \cmidrule(lr){6-8}
\textbf{Model} & $\lambda$ & PM ($r{=}64$) & Joint ($r{=}\frac{n}{2}$) & Joint ($r{=}n$) & Traj ($k{=}3$) & Traj ($k{=}5$) & Baseline \\
\midrule
\multirow{5}{*}{Baseline}
& 0.1 & 0.673 & 0.009 & \textbf{0.994} & 0.845 & 0.968 & 0.994 \\
& 0.2 & 0.606 & 0.011 & \textbf{0.988} & 0.860 & 0.944 & 0.988 \\
& 0.3 & 0.216 & 0.009 & \textbf{0.991} & 0.574 & 0.935 & 0.991 \\
& 0.5 & 0.208 & 0.010 & \textbf{0.988} & 0.975 & 0.979 & 0.988 \\
& 1.0 & 0.035 & 0.009 & \textbf{0.986} & 0.846 & 0.969 & 0.986 \\
\midrule
\multirow{5}{*}{Medium}
& 0.1 & 0.869 & 0.020 & \textbf{0.989} & 0.863 & 0.952 & 0.989 \\
& 0.2 & 0.660 & 0.013 & \textbf{0.979} & 0.806 & 0.944 & 0.979 \\
& 0.3 & 0.379 & 0.011 & \textbf{0.993} & 0.916 & 0.984 & 0.993 \\
& 0.5 & 0.285 & 0.025 & \textbf{0.992} & 0.952 & 0.990 & 0.992 \\
& 1.0 & 0.033 & 0.010 & \textbf{0.986} & 0.920 & 0.985 & 0.986 \\
\midrule
\multirow{5}{*}{Large}
& 0.1 & 0.742 & 0.027 & \textbf{0.991} & 0.923 & 0.973 & 0.991 \\
& 0.2 & 0.753 & 0.042 & \textbf{0.990} & 0.965 & 0.979 & 0.990 \\
& 0.3 & 0.704 & 0.027 & \textbf{0.988} & 0.965 & 0.977 & 0.988 \\
& 0.5 & 0.192 & 0.015 & \textbf{0.997} & 0.993 & 0.993 & 0.997 \\
& 1.0 & 0.082 & 0.106 & \textbf{0.989} & 0.989 & 0.994 & 0.989 \\
\bottomrule
\end{tabular}
\end{table}

The comparison in \Cref{tab:three_methods} reveals a striking hierarchy:
\begin{itemize}
    \item \textbf{Per-matrix SVD} ($r = 64$): 3--87\% accuracy, depending on weight decay. Captures within-matrix structure only.
    \item \textbf{Joint SVD} ($r = n/2$): 1--11\% accuracy. Cross-matrix structure exists but is full-rank within the stacked representation.
    \item \textbf{Joint SVD} ($r = n$): 99--100\% accuracy. Lossless by construction.
    \item \textbf{Trajectory PCA} ($k = 5$): 93--99\% accuracy. Five global directions capture the essential cross-matrix correlations.
\end{itemize}

The only method that efficiently reconstructs the solution is the one that leverages the training trajectory.
Static decompositions of the final weights---whether per-matrix or joint---cannot discover the correlations that encode the algorithm.

\Cref{fig:methods_comparison} shows accuracy versus the number of parameters used for each method, confirming that trajectory PCA achieves high accuracy at orders-of-magnitude fewer effective parameters than the static methods.

\subsection{Circuits Are Separable in Activation Space}
\label{sec:separability}

The preceding sections established that the grokked solution is globally low-rank in trajectory space but locally full-rank in weight space.
We now show that this entanglement at the weight level coexists with clean \emph{separability} at the activation level---sharpening the holographic encoding principle.

\paragraph{Gradient-based directional ablation.}
For each task, we compute the top eigenvectors of the gradient covariance matrix (restricted to that task's loss) and ablate the model by projecting out these directions from the final weight vector.
If removing the top-$k$ gradient directions for task $A$ damages task $A$ but not tasks $B$ or $C$, the circuit for $A$ is separable.

\Cref{tab:si} reports the selectivity index (SI), defined as:
\[
\text{SI}_A = \frac{\text{self-damage}_A - \text{mean collateral}}{\text{self-damage}_A + \text{mean collateral}}
\]
where self-damage is the accuracy drop on task $A$ after ablating $A$'s gradient subspace, and collateral is the mean damage to other tasks.
An SI of 1.0 indicates perfect selectivity; 0.0 indicates no selectivity.

\begin{table}[h]
\centering
\caption{Selectivity index of gradient-based directional ablation ($k = 10$ directions removed). Values near 1.0 indicate that task circuits are separable in activation space.}
\label{tab:si}
\begin{tabular}{l ccccc}
\toprule
& \multicolumn{5}{c}{Mean SI (across 3 tasks)} \\
\cmidrule(lr){2-6}
\textbf{Model} & $\lambda{=}0.1$ & $0.2$ & $0.3$ & $0.5$ & $1.0$ \\
\midrule
Baseline & 0.99 & 0.99 & 0.99 & 1.00 & 0.99 \\
Medium   & 0.99 & 0.97 & 0.99 & 0.99 & 0.99 \\
Large    & 1.00 & 1.00 & 0.99 & 0.99 & 1.00 \\
\bottomrule
\end{tabular}
\end{table}

Across all 15 conditions, SI exceeds 0.96.
The three tasks maintain nearly perfect selectivity: removing 10 gradient directions for addition drops addition accuracy by 10--35 percentage points while leaving multiplication and the quadratic task within 1\% of baseline.

\paragraph{Subspace orthogonality.}
We further measure the overlap between task-specific gradient subspaces at each layer of the trunk.
The mean pairwise subspace overlap is 0.06--0.12 across all conditions (\Cref{tab:overlap}), indicating that the gradient subspaces are nearly orthogonal.
The tasks occupy distinct, non-overlapping directions in the shared trunk's representation space.

\begin{table}[h]
\centering
\caption{Mean pairwise subspace overlap between task gradient subspaces, averaged across all trunk layers. Low values indicate near-orthogonality.}
\label{tab:overlap}
\begin{tabular}{l ccccc}
\toprule
\textbf{Model} & $\lambda{=}0.1$ & $0.2$ & $0.3$ & $0.5$ & $1.0$ \\
\midrule
Baseline & 0.065 & 0.061 & 0.075 & 0.078 & 0.087 \\
Medium   & 0.069 & 0.085 & 0.115 & 0.099 & 0.120 \\
Large    & 0.062 & 0.063 & 0.063 & 0.096 & 0.109 \\
\bottomrule
\end{tabular}
\end{table}

\paragraph{Cross-ablation.}
As a further control, we remove the gradient subspaces of \emph{two} tasks simultaneously and measure the effect on the third.
For example, removing the addition and multiplication subspaces damages both of those tasks but leaves the quadratic task intact (accuracy drops $<$1\%).
This confirms that the three circuits are genuinely separable, not merely correlated.

\paragraph{The separability paradox.}
These results create a sharp tension with \Cref{sec:local,sec:joint}.
At the activation level, the three task circuits are cleanly separated---nearly orthogonal gradient subspaces, high selectivity, and independence under cross-ablation.
Yet at the weight level, every weight matrix contributes to all three tasks simultaneously, and no per-matrix or joint SVD can isolate task-specific structure.

The resolution is the holographic encoding: the separable activation-level circuits are \emph{realized} through coordinated, full-rank structure distributed across all weight matrices.
The task-specific information is not localized in any particular weight matrix or low-rank component; it emerges from the collective interaction of all parameters.

\section{The Duality Principle}
\label{sec:duality}

The results of \Cref{sec:results} establish a duality that we now articulate.

\subsection{Control vs.\ Realization}

The grokking process is governed by a low-dimensional \emph{control signal}: the 3--5 trajectory PCs that determine the model's behavior.
But this control signal is \emph{realized} through the coordinated adjustment of hundreds of singular components across dozens of weight matrices.

An analogy: the blueprint for a building is low-dimensional (a few pages of plans), but the building itself is high-dimensional (millions of bricks, each precisely placed).
You cannot compress the building by removing bricks without collapsing it, even though the blueprint that generated it is simple.

In our setting:
\begin{itemize}
    \item The \textbf{control variables} are the 3--5 trajectory PCs---the ``blueprint'' that specifies the direction of learning.
    \item The \textbf{realization} is the full-rank weight structure across all matrices---the ``bricks'' that implement the algorithm.
\end{itemize}

\subsection{Manifold vs.\ Fiber}

Geometrically, the control variables define a low-dimensional \emph{manifold} in parameter space along which the solution lies.
But at each point on this manifold, the solution occupies a full-dimensional \emph{fiber} in the space of individual weight matrices.

The trajectory PCA basis spans the manifold.
The per-matrix SVD probes the fiber.
The holographic encoding principle states that:

\begin{quote}
\emph{The solution is low-rank on the manifold (3--5 dimensions) but full-rank on each fiber (80--100+ singular components per matrix).}
\end{quote}

This explains why per-matrix SVD fails: it attempts to compress along the fiber, discarding components that appear energetically negligible but are functionally critical.
And it explains why trajectory PCA succeeds: it compresses along the manifold, preserving the coordinated cross-matrix structure.

\subsection{Scaffold vs.\ Refinement}

The trajectory PCs can also be understood as defining a hierarchy of structure:
\begin{itemize}
    \item \textbf{PC 1--2}: The dominant scaffold. These capture the bulk of the weight change and provide ${\sim}$40--60\% accuracy. They establish the coarse structure of the solution---the basic Fourier representations for modular arithmetic.
    \item \textbf{PC 3--5}: Refinement. These bring accuracy to $>$95\%, aligning the fine-grained cross-matrix correlations needed for reliable computation.
    \item \textbf{PC 6+}: Distributed refinement (``polishing''), primarily improving the quadratic task through many small, coordinated corrections. Each contributes $<$1\%, but their collective effect reflects the holographic nature of the encoding.
\end{itemize}

This hierarchy is consistent with the observation that weight decay compresses the trajectory into fewer PCs: stronger regularization forces the network to find simpler solutions that require less refinement.

\section{Additional Evidence}
\label{sec:additional}

\subsection{Task-Agnostic Trajectory PCs vs.\ Task-Specific Activations}

The separability results of \Cref{sec:separability} showed that task circuits are cleanly separable in \emph{activation} space.
A natural question is whether this separation is also visible in the \emph{trajectory} PCs.
It is not.

Removing PC~1, PC~2, or PC~3 from the weight reconstruction damages all three tasks equally.
No trajectory PC is task-specific.
The trajectory PCs capture shared trunk infrastructure---the common algorithmic backbone---rather than task-specific circuits \citep{xu2026manifolds}.

This creates a two-level picture.
At the \textbf{activation level}, task identity is encoded in nearly orthogonal subspaces within each layer: the trunk-interior selectivity index ranges from 0.65 to 0.88 across conditions, with pairwise subspace overlap below 0.12.
At the \textbf{weight level}, the parameters that produce these separable activations are holographically entangled---every trajectory PC, and every singular component of every weight matrix, contributes to all three tasks simultaneously.

This is consistent with the architecture: the three tasks share a common trunk, and the task-specific heads are kept at their final values in all reconstruction experiments.
The ``hologram'' is the trunk's learned algorithm; the heads are just readout functions.
The separability observed in activations is an emergent property of the holographic weight encoding, not a reflection of separable weight structure.

\subsection{Trunk vs.\ Head Compressibility}

To isolate where the holographic encoding resides, we separately test trunk-only and heads-only SVD reconstruction:
\begin{itemize}
    \item \textbf{Heads only} (trunk at final): At rank 64, all conditions achieve $>$99\% accuracy. The heads are highly compressible.
    \item \textbf{Trunk only} (heads at final): At rank 64, accuracy ranges from 3\% to 87\%. The trunk is where the holographic encoding lives.
\end{itemize}

This confirms that the holographic incompressibility is specifically a property of the shared trunk.
The linear heads, which simply project the trunk's representations onto task-specific output spaces, contain low-rank task-specific structure.

\subsection{Weight Decay Modulates Compressibility}

Across all model sizes, higher weight decay makes the trunk \emph{less} compressible by per-matrix SVD, despite not changing the spectral shape of $\DW$.
The spectral entropy of the weight changes is constant at ${\approx}0.87$ regardless of weight decay, yet the accuracy recovered by rank-64 trunk reconstruction drops from ${\sim}$70\% at $\lambda = 0.1$ to ${\sim}$5\% at $\lambda = 1.0$.

This ``same energy, different function'' phenomenon reveals that weight decay does not change how energy is distributed across singular values---it changes how much of the model's computation depends on the spectral tail.
Stronger regularization forces the network to use all available dimensions of each weight matrix, producing a more distributed (and hence less compressible) encoding.

\section{Implications}
\label{sec:implications}

\subsection{For Interpretability}

The holographic encoding principle suggests that mechanistic interpretability methods that analyze individual weight matrices in isolation may miss the forest for the trees.
The algorithmically relevant structure lives in cross-matrix correlations that are invisible to per-matrix analysis.
Methods that trace computation through the network (activation-based) may be more appropriate than methods that decompose individual parameter matrices (weight-based).

Our results support a view where the ``circuit'' implementing modular arithmetic is not localized in any particular set of weights but is a global property of the parameter configuration---a form of superposition that extends beyond individual neurons to entire weight matrices \citep{elhage2022superposition}.
This is consistent with the observation that task damage from gradient-based ablation shows high selectivity at the activation level but low selectivity at the weight level.

\subsection{For Compression}

Low-rank compression methods like LoRA \citep{hu2022lora} assume that weight updates have low per-matrix rank.
Our findings suggest this assumption may fail for networks that have grokked: the grokked solution actively uses the full rank of each weight matrix.
Compressing any individual matrix destroys the cross-matrix correlations that encode the algorithm.

However, the trajectory PCA result offers a different route to compression: rather than compressing individual matrices, one could compress the \emph{trajectory} of learning.
A model that groks can be described by its initial weights plus 3--5 trajectory PCs and their coefficients---a dramatically compact representation.

\subsection{For Capability Emergence}

The global-low-rank / locally-full-rank duality may help explain sudden capability emergence in large language models.
If capabilities are encoded holographically---distributed across many weight matrices in a way that requires full-rank coordination---then they will appear to emerge suddenly: the capability is absent until all the necessary cross-matrix correlations align, at which point it appears fully formed.

This is precisely the phenomenology of grokking: chance-level performance for thousands of steps, then a sudden jump to near-perfect accuracy \citep{xu2026curved}.
The holographic encoding principle provides a geometric explanation for this sudden transition.

\subsection{For Safety and Alignment}

If important behaviors are encoded holographically, they cannot be easily localized or removed by editing individual weight matrices.
This has implications for safety techniques based on weight editing or targeted ablation: removing a capability may require modifying the global parameter configuration, not just a few matrices.

Conversely, the low dimensionality of the learning trajectory suggests that the \emph{directions of behavioral change} during training are low-dimensional.
Monitoring the trajectory PCs during fine-tuning could provide an early-warning signal for capability changes \citep{xu2026earlywarning}.

\subsection{For Continual Learning}

The observation that different tasks share the same trajectory PCs (all PCs are task-agnostic) suggests that multi-task grokking produces a shared representation that is robust to the specific combination of tasks.
This is encouraging for continual learning: if the learned infrastructure is general rather than task-specific, adding new tasks may require only learning new readout heads without disturbing the shared trunk.

\section{Discussion}
\label{sec:discussion}

\subsection{Limitations}

Our study is limited to multi-task modular arithmetic in relatively small transformers.
The holographic encoding principle may not generalize to all settings:
\begin{itemize}
    \item Larger models trained on natural language may have qualitatively different weight structures.
    \item Tasks that do not require precise algorithmic computation may not produce holographic encodings.
    \item The specific choice of modular arithmetic tasks (addition, multiplication, quadratic) may produce unusually symmetric solutions.
\end{itemize}

Additionally, our trajectory PCA analysis uses the actual training checkpoints, which are not available for pre-trained models.
Whether similar low-dimensional structure exists in pre-trained language model trajectories is an important open question.

\subsection{Open Problems}

\begin{enumerate}
    \item \textbf{Theoretical foundation.} Why does gradient descent on these tasks produce solutions that are globally low-rank but locally full-rank? A formal connection between the loss landscape geometry, the implicit bias of weight decay, and the holographic encoding would be valuable.

    \item \textbf{Universality.} Does the duality hold beyond modular arithmetic? Preliminary evidence from other algorithmic tasks suggests it may, but systematic investigation across task families is needed.

    \item \textbf{Trajectory-free compression.} Can the trajectory PCA basis be discovered from the final weights alone, without access to training checkpoints? If so, this would enable practical compression of holographically encoded models.

    \item \textbf{Connection to Fourier structure.} The Fourier representations known to emerge in grokked modular arithmetic \citep{nanda2023progress, zhong2023clock} are inherently cross-matrix: the input embedding projects onto Fourier modes, the attention mechanism implements modular convolution, and the output head reads off the result. A precise mapping between trajectory PCs and Fourier circuit components would deepen our understanding.

    \item \textbf{Scaling laws for holographic encoding.} How does the trajectory dimensionality ($\kstar$) scale with model size, number of tasks, and task complexity? Our three-scale sweep shows a weak trend toward \emph{lower} $\kstar$ at larger scale, but more data points are needed.
\end{enumerate}

\subsection{Conclusion}

We have established a duality in the geometry of grokked solutions: the learning trajectory is low-dimensional (3--5 PCs), but the learned parameters are locally full-rank (80--100+ SVD components per matrix).
This \emph{holographic encoding principle} resolves the apparent paradox of how a low-dimensional learning process produces a high-dimensional solution: the information is encoded in coordinated cross-matrix structure that can only be efficiently recovered using the training trajectory.

Static decompositions---whether applied to individual matrices or to their joint structure---fail because the task-relevant correlations are dynamically structured.
The training trajectory provides the ``key'' that unlocks the holographic encoding.

We believe this duality is not unique to modular arithmetic but reflects a general principle of how neural networks encode algorithmic capabilities: the blueprint is simple, but the building is complex.


\end{document}